\pdfoutput=1
\documentclass[eat,twocolumn]{jmlr}

\usepackage{longtable}

\usepackage{longtable}% for long tables

\usepackage{booktabs}
\usepackage[load-configurations=version-1]{siunitx} % newer version
 \usepackage{xcolor}
\usepackage{array}

\def\appxbar[#1]#2{
    {#1 \color{purple}\rule{#2in}{8pt}}}

\theorembodyfont{\upshape}
\theoremheaderfont{\scshape}
\theorempostheader{:}
\theoremsep{\newline}

\jmlrvolume{}
\firstpageno{1}
\jmlryear{2022}
\jmlrworkshop{Machine Learning for Health (ML4H) 2022}

\title[Using Open-Ended Stressors to Predict Depressive Symptoms]{Using Open-Ended Stressor Responses to Predict Depressive Symptoms across Demographics}

  \author{\Name{Carlos Aguirre} \Email{caguirre@cs.jhu.edu} \\
  \Name{Mark Dredze}\\
  \addr Johns Hopkins University \\
  \AND \Name{Philip Resnik} \\ \addr University of Maryland}

\begin{document}

\maketitle

\begin{abstract}
Stressors are related to depression, but this relationship is complex.
We investigate  the relationship between open-ended text responses about stressors and depressive symptoms across gender and racial/ethnic groups.
First, we use topic models and other NLP tools to find thematic and vocabulary differences when reporting stressors across demographic groups. 
We train language models using self-reported stressors to predict depressive symptoms, finding a relationship between stressors and depression.
Finally, we find that differences in stressors translate to downstream performance differences across demographic groups.

\end{abstract}

\section{Introduction}
\label{sec:intro}

Despite depression being one of the most common mental health disorders, underrepresented communities continue to be under-diagnosed and under-treated \citep{sorkin2011underdiagnosed, stockdale2008racial}.
Further, there is a well documented difference in depression prevalence in gender \citep{CDC} and racial/ethnic groups \citep{barnes2013racial, woodward2013major}.
One theory for the differences in depression prevalence, diagnosis and treatment across demographics groups is \textit{Cultural Relativity}, which attributes these variations to differential life events, vulnerability and symptoms \citep{whaley1997ethnicity,moazen2016depressive, assari2016ethnic}.
Given that stressors have been shown to differ across demographics \citep{pearlin2005stress}, stress has been thought to play a role in depression differences by demographics \citep{nolen2009gender}.
% Following this theory, prior work has studied the relationship of stressors and depression.
The hypothesis is that certain demographic groups, e.g. women or African Americans, have greater exposure and/or are more vulnerable to stressors, which has been associated with depressed mood and general anxiety \citep{bowleg2020negative, fedina2018police, kwate2015cross, marchand2016gendered, williams1997racial}.
However, recent studies have stressed possible biases in the analysis of stressors studies, e.g. standardized stressor questionnaires lack the power or specificity of interview or open-ended measures \citep{acharya2018college}.

We take advantage of an unusual data collection effort that provides access to clinically validated scales of depressive symptoms as well as open-ended survey responses about stressors, to analyze the relationship of stressors and depression across gender and racial/ethnic groups.
Open-ended survey questions may be particularly effective at studying the complex relationship of stressors and depression across demographics \citep{weger2022there}.
Our hypothesis is that racial/ethnic and gender groups share different stressors, and these responses are not equally predictive of their actual mental health state.
To analyze differences in text across demographics, we automatically analyze survey text responses using methods based on vocabulary distributions, such as log-odds-ratios and Linguistic Inquiry and Word Count \citep[LIWC]{pennebaker2015development} distributions, as well as topics discovered by a structural topic model, which allows covariates to affect topical prevalence and topical content \citep{roberts2013structural}.
We also train language models to study the relationship of stressors and depression across demographics.
In summary, our contributions are the following:
\begin{itemize}
    \setlength\itemsep{.01em}
    \item We analyze differences in stressors across gender and racial/ethnic groups using open-ended text responses
    \item We contextualize the relationship of stressors and depression by using text responses of stressors as predictors for depression
    \item We analyze the performance of these predictive models across gender and racial/ethnic groups
\end{itemize}

\section{Background}
\label{sec:background}

Stressors have a complex relationship with depression \citep{acharya2018college}.
For instance, while studies have shown that stressful events are associated with episodes of depression \citep{wike2014discrimination, hammen2005stress, kessler1997effects, mazure1998life}, the neurobiological response to stress does not always lead to depressive symptoms \citep{van2004stress, bonde2008psychosocial}.
Therefore, while determining the exact nature of the relationship between stressors and depression may lead to more effective treatment \citep{van2004stress}, we do not expect high performance from models that only utilize stressors to predict depression. 

Relevant to our study, stressors and stress levels are known to vary across demographics, such as gender \citep{acharya2018college, parker2010gender} and race \citep{alang2021police, pearlin1997forms}.
Stressors related to changes in social activities, eating, workload, education performance, living environments, parental relations, and financial strain are more prevalent in women  \citep{acharya2018college, lewis2015gender, cherenack2022puberty, elliott2001gender}.
Black college students experience additional stress related to being a racial minority in predominantly White settings \citep{ancis2000student}.
Other stressors, such as police encounters, have been linked with Latinx and African American populations compared to White populations \citep{alang2021police}.
The dichotomy of stressors is usually studied between majority White populations and a minority racial group, however, some work suggests that race-related stressors may be different across non-White populations, e.g. African American students self-report higher levels of race-related stress compared to Chicano and Filipino counterparts \citep{baker1988relationship}.

\vspace{-.4em}
\section{Methods}
\label{sec:methods}
We present the dataset used in our study (\sectionref{sec:methods-data}), our procedure for analyzing stressors (\sectionref{sec:methods-stressors}), and the details of the prediction task (\sectionref{sec:methods-prediction}.)
\vspace{-.5em}
\subsection{Data}
\label{sec:methods-data}
We use the UMD-ODH dataset \citep{kelly2020blinded, kelly2021can}, which includes self-reported demographics and clinically validated mental health scales for depressive symptoms, QIDS-sr16 \citep{rush200316}, and psychosis, CAPE-15 \citep{capra2017current}, as well as an open-ended question about stressors: \textit{Describe the biggest source of stress in your life at the moment. What things have you done to deal with it?}

Participants were allowed to respond in follow-up surveys, with $\sim$47\% of subjects doing 2 or more survey responses and an average $\sim$3.3 surveys per subject.
We select the survey responses that have demographic data available, and remove duplicates and responses with less than 3 tokens resulting in $2607$ survey responses.
We used the CLD3\footnote{https://github.com/google/cld3} language identification tool and Google Translate\footnote{https://github.com/ssut/py-googletrans} to find and translate non-English responses ($< 5\%$ Spanish).
We obtain the depressive symptom labels from QIDS-sr16 cutoffs: 7 for minor depressive symptoms and 9 for major depressive symptoms, following prior work \citep{sung2013screening}.
\tableref{tab:data-stats} shows the dataset statistics for depressive symptoms and demographics.

\vspace{-.3em}
\subsection{Stressors Analysis}
\label{sec:methods-stressors}
We analyze open-ended text responses on stressors using both vocabulary and topic model methods.
As we summarized in \sectionref{sec:background}, there is evidence that stressors differ across demographics.
Therefore, we use these methods to compare distributions of two groups: pair-wise comparisons for gender and racial/ethnic groups.

\textbf{Vocabulary Analysis.} 
To analyze the vocabulary distributions of the survey responses we employed the log-odds-ratio, informative Dirichlet prior\footnote{COCA English corpus is the prior \citep{davies2008corpus}}, which is often used to study vocabulary differences between groups \citep{monroe2008fightin}.
We utilized the Spacy English tokenizer \citep{spacy2}, and remove punctuation and digits.
Tokens that were not used in at least 5 different surveys were removed from the vocabulary.
Additionally, in order to avoid subject-bias in the analysis, we experimented with using either log of frequencies (less restrictive) or bag-of-words (more restrictive) at the subject level to restrict the impact each individual respondent had on the distribution, finding that log of frequencies yielded a good balance.
In addition to vocabulary, we also used categories from the LIWC dictionary, which has been used in other mental health tasks \citep{aguirre-dredze-2021-qualitative}.

\textbf{Topic Models.}
We trained a structural topic model (STM) to compare the topic distributions across covariates \citep{roberts2013structural}.
The covariates for the analysis are the gender and racial/ethnic labels, and the depressive symptom labels based on QIDS-sr19 cutoffs.
We also add a COVID-19 covariate, as we observe differences in populations before and after the COVID-19 pandemic.\footnote{COVID-19 population analysis in \appendixref{apd:covid-19}}
An author of the paper and two non-author annotators agreed on topic cohesion (whether a topic has a name or not) and names using the top 30 tokens and text responses ranked by topic proportion; Fleiss' kappa showed that there was fair agreement between the annotators, $\kappa=0.357$.
\figureref{fig:black_white_pretty} show the effect of the race covariate (Black vs White participants) on the final topics annotated.

\subsection{Prediction Task}
\label{sec:methods-prediction}
We formulate the prediction task as a binary classification task based on the QIDS-sr19 cutoff scores by joining the Minor and Major Depressive symptom surveys into one category.\footnote{We also experimented with control vs Major Depressive symptoms as labels in \appendixref{apd:mdd-task}}
We performed a random train-test split (90-10\%) ensuring that surveys from the same subject were not shared across splits.
We train both linear and neural models.

\textit{Linear.} 
Inspired by prior work predicting depression using social media, we utilize TF-IDF, 50-dimensional LDA topic distribution LDA \citep{blei2003latent}, 200-dimensional mean-pooled vector of GloVe embeddings \cite{pennington2014glove} and LIWC categories as features for our models \citep{aguirre2021gender}.
We experimented with $l_2$-regularized Logistic Regression and SGD Classifiers.
See \appendixref{apd:model-specs} for more details, e.g. hyperparameter search space.

\textit{Neural.}
We experimented with fine-tuning MentalBERT \citep{ji2022mentalbert}, an example of a recently introduced large language model for mental health classification tasks.
In order to contextualize the performance due to pretraining on mental health data, we also finetuned BERT \citep{devlin-2018-bert} and ClinicalBERT \citep{alsentzer-etal-2019-publicly}.
For each task, we fine-tuned all layers of the models in addition to the classification head. 
Following \citet{barbieri-etal-2020-tweeteval}, we performed a parameter search over learning rate values of $1e^{-3}$, $1e^{-4}$, $1e^{-5}$, and $1e^{-6}$, with a batch size of $32$ for at least $20$ epochs.

\begin{figure}[t]
\floatconts
  {fig:black_white_pretty}
  {\caption{Stressor topics more related to Black vs White participants.}}
  {\includegraphics[width=1.0\linewidth]{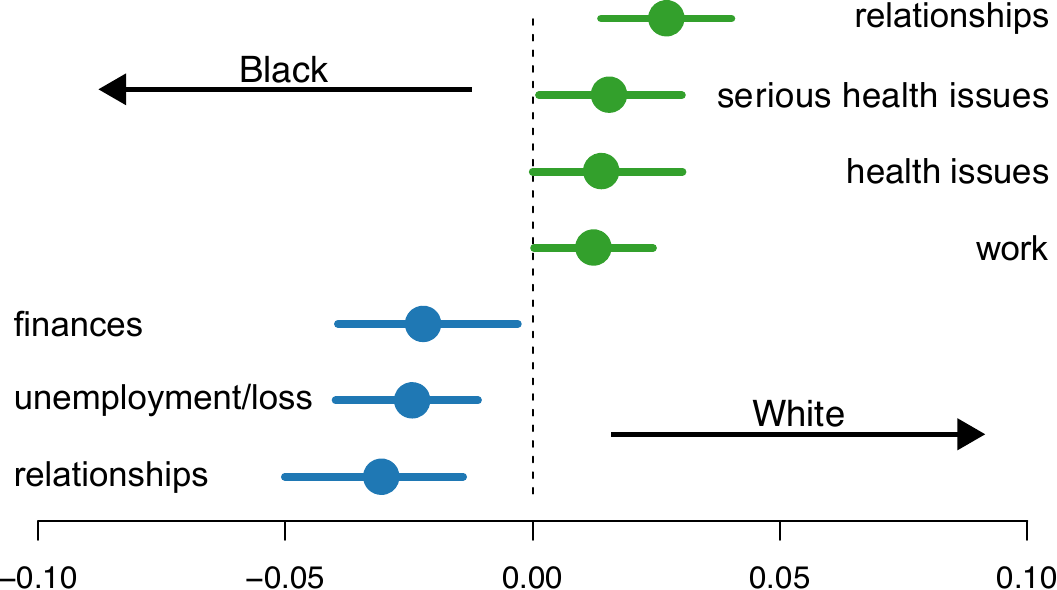}}
\end{figure}

\textbf{Fairness Metric.}
Language models are known to suffer from biases associated with demographics \citep{aguirre2021gender}. 
Furthermore, models may be fair across a single demographic axis, but biases for \textit{intersectional} groups (known as \textit{Fairness Gerrymandering} \citep{kearns2018preventing}.)
We check the performance of our models across intersectional groups.
Intersectional fairness has been the focus of recent work, either adapting single axis fairness definitions to intersectional groups \citep{ghosh2021characterizing} or creating new definitions explicitly for intersectional settings \citep{kearns2018preventing}.
We utilize an empirical estimation of an $\epsilon-$Differential Fairness \citep{foulds2020intersectional} definition of Equalized Odds adapted from \cite{morina2019auditing}: $\epsilon$-Differential Equalized Odds ($\epsilon$-D).\footnote{See \appendixref{apd:fairness-definition} for the detailed definition.}

\vspace{-1em}
\section{Results \& Analysis}

\subsection{Stressors Analysis}
\label{sec:results-stressors}

We report the analyses of the stressors reported in the survey responses. \figureref{fig:black_white_pretty} shows an example of the results the topic model when comparing Black and White individuals in the dataset;
\appendixref{apd:stressors-figures} shows the rest of the topic model and log-odds-ratio, Informative Dirichlet prior pair-wise comparisons across the demographics.

\textbf{Gender.}
We find stressors related to \textit{health} (vocab: COVID-19, pandemic, exercising; liwc: health), \textit{finances} (vocab: finances; liwc: money, reward), \textit{relationships} (vocab: boyfriend; liwc: family, home) and the \textit{self} (vocab: I'm, my, me, myself; liwc: 1st-person-pronoun) more likely in women than men population.
Stressors related to \textit{social interactions} (vocab: he, his, their, she, we; liwc: 2nd\&3rd-person-pronoun) more likely in men than women.
Stressors related to \textit{education} (vocab: schoolwork, homework, finals) were related to both.

Similar to prior work, we find stressors related to finances, relationships and health are more prevalent for women than men.
However, we find that stressors related to education are prevalent in both groups, contradicting prior work \citep{acharya2018college}.

\textbf{Race/Ethnicity.}
In the African American group, text responses related to \textit{community} (vocab: programs[governmental], neighbors) and \textit{religion} (liwc: relig) were more likely than other groups.
In the Asian American group, stressors related to \textit{scheduling} (vocab: scheduling, managing, organized, balancing, time; liwc: time) were more likely than other groups.
In the Hispanic/Latinx group, stressors related to \textit{government} (vocab: government, state; liwc: social), \textit{death} (vocab: loss[family member/friend]), and \textit{basic needs} (vocab: basic[needs], daily) were more likely than other groups.
And finally, in the White American group, text responses related to \textit{family} (liwc: family, home) and \textit{coping mechanisms} (vocab: therapist, relax, yoga, cope) were more likely than other groups.
Stressors related to \textit{education} were related to Hispanic/Latinx, Black and Asian groups;
stressors related to \textit{health} were more related with Hispanic/Latinx, Black and White groups; and,
stressors related to \textit{finances} were related to all groups.

Consistent with prior work, we found stressors related to poverty, e.g. \textit{community} in African Americans and \textit{government} in Hispanic/Latinx, related more with marginalized populations, and
that \textit{religion}, more prevalent with African Americans, may be a coping mechanism \citep{strawbridge1998religiosity}.

\vspace{-1.25em}
\subsection{Prediction Task}
\label{sec:results-prediction}

\begin{figure*}[t]

\floatconts
  {fig:model-performance}
  {\caption{F1 macro-averaged scores of test set of models by gender and racial/ethnic groups.}}
  {\includegraphics[width=1\linewidth]{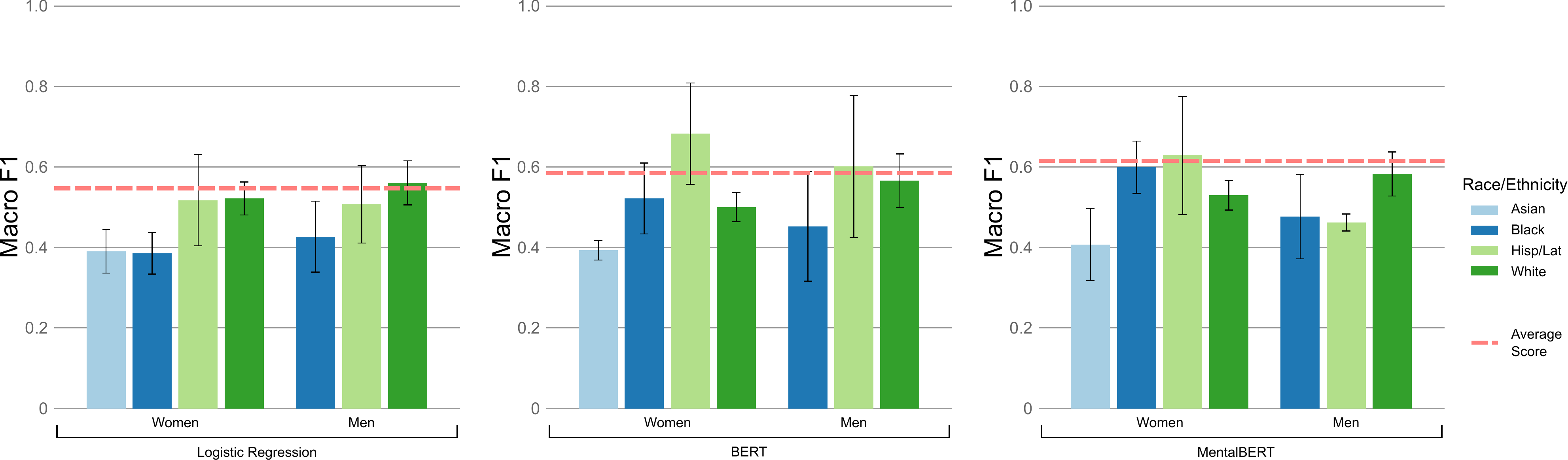}}
\end{figure*}

We performed a hyperparameter search using 10-fold cross validation and report the best mean and standard deviation macro-averaged F1 score of the validation sets for each model, as well as the corresponding test set performance in \tableref{tab:results-prediction-mdd-task}.

Results suggest that the neural models outperform linear models.
This performance difference is expected since the neural models have been pre-trained with much more data and have been shown to outperform linear models in similar tasks \citep{8998086}.
Still, neural models have drawback, e.g. lack of interpretability, more compute resources.

Comparing the neural models, we expected MentalBERT to outperform the other variants of BERT, as it was further trained on mental health related text and was shown to outperform these models on similar tasks based on social media.
However, we observe minimal performance differences between MentalBERT, BERT and ClinicalBERT, suggesting that the mental health pre-training performance gain may not generalize to other mental health tasks or domains.

\begin{table}[t]
\begin{small}

\floatconts
  {tab:results-fairness}
  {\caption{$\epsilon$-D of the combined 10-fold cross validation models for both the validation and test sets}}
  {\begin{tabular}{rcccc}
  \toprule
                    & Validation & Test \\

\cmidrule(lr){2-2}  \cmidrule(lr){3-3}  
Logistic Regression & \textbf{0.20}  & \textit{1.61} \\
SGD Classifier      & 0.34           & \textbf{0.76} \\
BERT                & 0.33           & 0.81          \\
ClinicalBERT        & 0.27           & 0.96          \\
MentalBERT          & \textit{0.37}  & 1.10          \\
\bottomrule
\end{tabular}}
\end{small}
\end{table}

Additionally, we investigate the performance difference of the models across demographics, as measured by $\epsilon$-Differential Equal Odds ($\epsilon$-D), by combining the predictions of the resulting models of the 10-fold cross validation. 
The results in \tableref{tab:results-fairness} suggest that the linear models tend to be fairer compared to the neural models, however, linear models perform closer to random performance which suggest that the performance is close to random for all demographic sub-groups.
Comparing the neural models, there doesn't seem to be a clear trend, except that MentalBERT seems to be the most unfair neural model.

The fairness and performance results suggest that there is a performance-fairness balance, where higher performing models tend to be less fair.
A difference compared with prior work is that the low fairness scores are not between majority and minority groups (e.g. often between White men and African American women), but actually from two underrepresented groups: Latinx (higher performance) and Asian American (lower performance), as seen in \figureref{fig:model-performance}.
This performance disparity may also be due to the relationship between stressors and depression across demographics, as some subgroups' stressors may be more related than others, as explored in \sectionref{sec:background}.

\vspace{-1.2em}
\section{Conclusion \& Future Work}

We studied the use of open-ended text responses about stressors to analyze the relationship between stressors and depression across gender and racial/ethnic groups.
We find that stressors vary across demographics.
Further, we train language models on the stressors responses to predict depressive symptoms, finding a relation between stressors and depression.
However, these stressors are not equally predictive of depressive symptoms across demographics, as we find a trade-off between overall model performance and fairness across demographics.

Future work may investigate the use of demographic specific cut-off scores for QIDS-sr16 as there is evidence these may be correlated with these demographics \citep{lamoureux2010using}.
Additionally, future annotation efforts on this dataset or others may enable studying the relationship between stressors and coping mechanism, providing more structure or increasing the size of the data that predictive models could use to increase the classification performance.
Further, while our study does not include racial/ethnic and gender small minorities due to insufficient sample size, future work could study these groups which often experience higher levels of stress which has been linked with elevated risk of depressive symptoms or major depressive disorder episodes \citep{hoy2016lesbian}.

This study makes a step towards understanding the relationship between stressors and depression contextualized by gender and racial/ethnic demographics, and could be expanded by efforts that seek to provide personalized care.
However, we must proceed with caution when seeking to generalize the relationship between stressors and depression across demographics.

\vspace{-1em}
\acks{We would like to acknowledge Rosalyn Shin and Elizabeth Salesky for annotating the topic model results. 
This work was supported in part by the National Science Foundation under award 2124270.
}

\bibliography{main}

\appendix
\onecolumn
\renewcommand{\thepage}{}

\section{Data Statistics}
\label{apd:data-stats}

\begin{table}[t]
\floatconts
{tab:data-stats}
  {\caption{UMD-ODH data statistics used in this study. Surveys were filtered based on open-ended text responses.}}
  {\begin{tabular}{lllll}

  \toprule
          &                            \multicolumn{2}{r}{\# surveys} & \%   &  \\
\midrule
\multicolumn{2}{l}{Total number}        & 2607       &      &  \\
\multicolumn{2}{l}{Gender}     &                            &            &        \\
          & women                     & 1854       & 71.1 &  \\
          & men                       & 659        & 25.3 &  \\
          & other                      & 94         & 3.6  &  \\
\multicolumn{2}{l}{Race/Ethnicity}      &            &      &  \\
          & Caucasian                  & 1761       & 67.5 &  \\
          & African American           & 221        & 8.5  &  \\
          & Asian American             & 246        & 9.4  &  \\
          & Latinx/Hispanic            & 277        & 10.6 &  \\
          & other                      & 102        & 3.9  &  \\
\multicolumn{2}{l}{Depressive Symptoms} &            &      &  \\
          & No Symptoms                & 1012       & 38.8 &  \\
          & Minor Dep. Symp. & 531        & 20.4 &  \\
          & Major Dep. Symp.  & 1064       & 40.8 &  \\
\multicolumn{2}{l}{COVID-19}   &                            &            &       \\
          & Before Pandemic            & 890        & 34.1 &  \\
          & During Pandemic            & 1717       & 65.9 &  \\
Age        &                            &            &      &  \\
          & 18-29                      & 1084       & 41.6 &  \\
          & 30-59                      & 1409       & 54.0 &  \\
          & \textgreater{}=60          & 114        & 4.4  & \\
          \bottomrule
\end{tabular}}
\end{table}

\tableref{tab:data-stats} shows the dataset statistics for depressive symptoms and demographics.
Surveys were filtered based on open-ended text responses as described in \sectionref{sec:methods-data}.

\newpage

\section{Covid-19 Analysis}
\label{apd:covid-19}

Prior work suggest that the prevalence for depression symptoms in the US was more than 3-fold higher during COVID-19 compared to before \citep{ettman2020prevalence}, and that there is a relation between stressors before and during the pandemic \citep{ettman2021low}.
To contextualize this effect in our dataset, we ask the following population questions: Are the depression rates before and during pandemic different across demographic groups? Do the demographic populations (race/ethnicity and gender) differ before and during the pandemic? 
We perform statistical tests on the number of surveys and report the results in this section.

We first compared the depression rates in the surveys submitted before and during the COVID-19 pandemic.
We performed 100 trials, where each trial sampled surveys with replacement for each demographic group and recording the resulting depression rate.
A pooled t-test revealed a significant difference in the depression rates of the surveys \textit{before} ($M=.56$, $SD=.028$) and \textit{during} the COVID-19 pandemic ($M=0.83$, $SD=.015$),  $t(198)=83.9$, $p<.001$.
This is also true for each demographic group; the tests results are reported in \appendixref{apd:pandemic-depression-results}.
These findings suggest that the rate of depression in the surveys was significantly higher during the COVID-19 pandemic than before.

Next, we compared the demographic populations of the surveys submitted before and during the COVID-19 pandemic.
We constructed contingency tables for each demographic axis: gender and racial/ethnic groups.
A $\chi^2$-test revealed a significant difference in racial/ethnic distributions for surveys before and during the COVID-19 pandemic but not for gender, suggesting that there are more surveys of Hispanic/Latino, Black and Asian respondents during the pandemic than before, but the gender populations did not change (see \appendixref{apd:pandemic-demographic-results} for more details).

Together, these results suggest that the population characteristics change based on COVID-19 pandemic both at the demographic level and at the depression rate level.
A likely reason may be changes in the recruitment strategy of participants, as the dataset creators reported multiple recruitment efforts both before and during the pandemic.
Other studies have found that depression rates increased during the COVID-19 pandemic \citep{ettman2020prevalence}.
Overall, these changes mean that we must control for the pandemic in our experiments to avoid probable confoundings.

\newpage
\section{Major Depressive Symptom Task}
\label{apd:mdd-task}

We formulate the prediction task as a binary classification task based on the QIDS-sr19 cutoff scores by joining the Minor and Major Depressive symptom surveys into one category in \sectionref{sec:methods-prediction}.
Here, we conduct experiments with a binary classification task where we only consider control vs Major Depressive symptoms (excluding Minor Depressive symptom surveys).
Similar to \sectionref{sec:methods-prediction}, we performed a random train-test split (90-10\%) ensuring that surveys from the same subject were not shared across splits, and the linear and neural model setup is the same.

\begin{table*}[t]
\floatconts
  {tab:results-prediction-mdd-task}
  {\caption{Mean ($\mu$) and standard deviation ($\sigma^2$) of macro-averaged F1 scores of 10-fold cross validation models of the validation and test sets}}
  {\begin{tabular}{rllllllll}
  \toprule
                    & \multicolumn{4}{c}{1. Depressive Symptoms}                   & \multicolumn{4}{c}{2. Major Depressive Symptoms}             \\
\cmidrule(lr){2-5}  \cmidrule(lr){6-9}
                    & \multicolumn{2}{c}{Validation} & \multicolumn{2}{c}{Test} & \multicolumn{2}{c}{Validation} & \multicolumn{2}{c}{Test} \\
\cmidrule(lr){2-3}  \cmidrule(lr){4-5} \cmidrule(lr){6-7}  \cmidrule(lr){8-9} 
Model               & $\mu$         & $\sigma^2$     & $\mu$      & $\sigma^2$  & $\mu$         & $\sigma^2$     & $\mu$      & $\sigma^2$         \\
\cmidrule(lr){2-2} \cmidrule(lr){3-3}  \cmidrule(lr){4-4} \cmidrule(lr){5-5} \cmidrule(lr){6-6} \cmidrule(lr){7-7}  \cmidrule(lr){8-8} \cmidrule(lr){9-9}
Logistic Regression & 0.53          & 0.051          & 0.55       & 0.031       & 0.53          & 0.033          & 0.60       & 0.033       \\
SGD Classifier      & 0.53          & 0.037          & 0.56       & 0.019       & 0.53          & 0.041          & 0.56       & 0.045       \\
BERT                & 0.54          & 0.063          & 0.59       & 0.030       & \textbf{0.60} & 0.057          & 0.60       & 0.031       \\
ClinicalBERT        & 0.52          & 0.058          & 0.61       & 0.027       & \textbf{0.60} & 0.050          & \textbf{0.62}       & 0.033       \\
MentalBERT          & \textbf{0.55} & 0.051          & \textbf{0.62}& 0.029     & 0.59          & 0.069          & \textbf{0.62}       & 0.035      \\

\bottomrule
\end{tabular}}
\end{table*}

\begin{table}[t]
\floatconts
  {tab:results-fairness-mdd-task}
  {\caption{$\epsilon$-Differential Equalized Odds ($\epsilon$-D) of the combined 10-fold cross validation models for both the validation and test sets}}
  {\begin{tabular}{rcccc}
  \toprule
                    & \multicolumn{2}{c}{Depressive Symptoms}                   & \multicolumn{2}{c}{Major Depressive Symptoms}             \\

                    & Validation & Test & Validation & Test \\

\cmidrule(lr){2-2}  \cmidrule(lr){3-3} \cmidrule(lr){4-4} \cmidrule(lr){5-5} 
Logistic Regression & \textbf{0.20}  & \textit{1.61} & 0.61           & 1.11          \\
SGD Classifier      & 0.34           & \textbf{0.76} & \textbf{0.30}  & \textbf{1.07} \\
BERT                & 0.33           & 0.81          & 0.63           & 2.09          \\
ClinicalBERT        & 0.27           & 0.96          & 0.64           & 1.34          \\
MentalBERT          & \textit{0.37}  & 1.10          & \textit{0.72}  & \textit{2.85} \\
\bottomrule
\end{tabular}}
\end{table}

We performed a hyperparameter search using 10-fold cross validation and report the best mean and standard deviation macro-averaged F1 score of the validation sets for each model, as well as the corresponding test set performance in \tableref{tab:results-prediction-mdd-task} (we include results in \sectionref{sec:results-prediction} for ease of comparison).
Overall, models trained in Task 2 tend to have higher F1 scores compared to Task 1,
which is surprising given that Task 2 has less data than Task 1.
Perhaps Task 2 is easier, as the less obvious Minor Depressive symptoms are removed, making a clearer distinction between responses of subjects with depressive symptoms and control.

We investigate the performance difference of the models across demographics.
We combined the predictions of the resulting models of the 10-fold cross validation and calculated fairness metrics based on our intersectional definition: $\epsilon$-Differential Equal Odds ($\epsilon$-D).
The fairness results in \tableref{tab:results-fairness-mdd-task} suggest that the linear models tend to be more fair compared to the neural models, although it varies depending on the dataset and task.
Comparing the neural models, there doesn't seem to be a clear trend, except in Task 2, where MentalBERT seems to be the most unfair model.

\newpage
\section{Model Specifications}
\label{apd:model-specs}

\textbf{Tokenization.}
Raw text from the survey responses was tokenized using a modified version of the Twokenizer \citep{o2010tweetmotif}. 
English contractions, hyphenated and slashed terms were expanded, while specific numeric values were deleted. 
As pronoun usage tends to differ in individuals living with depression \citep{vedula2017emotional}, we removed any English pronouns from our stop word set (English Stop Words from nltk.org). 
Case was standardized across all tokens.

\textbf{Features}. 
The bag-of-words representation of each survey is then used to generate the following additional feature dimensions: a 50-dimensional LDA topic distribution \citep{blei2003latent}, a 64-dimensional LIWC category distribution \citep{pennebaker2015development}, and a 200-dimensional mean-pooled vector of GloVe embeddings \citep{pennington2014glove}. 
The reduced bag-of-words representation is transformed using TF-IDF weighting \citep{ramos2003using}.\footnote{All data-specific feature transformations (e.g. LDA, TF-IDF) are learned without access to development or test data.}

\textbf{Hyperparameter Selection.} 
Each model is trained using a hyperparameter grid search over the regularization strength \{1e-3, 1e-2, 1-e1, 1, 10, 100, 1e3, 1e4, 1e5\}, class weighting \{None, Balanced\}, and feature set standardization \{On, Off\}, as well as the model-specific hyperparameters shown in \tableref{tab:hyperparameter-linear}.

\begin{table}[t]

\floatconts
  {tab:hyperparameter-linear}
  {\caption{Model specific hyperparameter search space}}
  {\begin{tabular}{rcl}
  \toprule
  Model & Parameter & Space \\
  \midrule
Logistic Regression & C  & 1000, 100, 10, 1, .1, .01, .001 \\
                    & solver & liblinear, lbfgs \\
SGD Classifier      & alpha  & 1e-5, 1e-4, 1e-3, 1e-2 \\
                    & epsilon & .001, .01, .1 \\
\bottomrule
\end{tabular}}
\end{table}

Hyperparameters were selected to maximize held-out F1 score within a 10\%-sied held-out split of the training data (10-fold cross validation within the training data).
We used the Ray[tune]\footnote{https://github.com/ray-project/tune-sklearn} library for the hyperparameter search.

\newpage
\section{Fairness Definition}
\label{apd:fairness-definition}

To choose which fairness definition is important for these models, we employ likely use cases, such as a screening tool to prioritize access to care for possible patients in need.
In such case, equal access to care across demographics, which can be modeled by equal true positive rate (TPR), is most important; this is often referred to as \textit{Equal Opportunity} \citep{hardt2016equality}.
However, this definition would reward high recall models without regard to precision, which could create situations in finite resource systems (e.g. healthcare) where resources are invested only in the majority community.
Instead, we use the more restrictive \textit{Equalized Odds} intersectional definitions, that also penalize low specificity models \citep{hardt2016equality}:

Inspired by the Empirical estimation of $\epsilon-$Differential Fairness \citep{foulds2020intersectional} and an Equalized Odds adaptation of $\epsilon-$Differential Fairness \citep{morina2019auditing}, we define the Empirical estimation of the $\epsilon-$Differential Equalized Odds ($\epsilon-$D):
Assuming discrete outcomes, $P_{Data}(y|s)=\frac{N_{y,s}}{N_s}$, where $N_{y,s}$, $N_s$ are empirical counts of their subscripted values in the dataset, and $A$ is a set of all subgroups of protected attributes. 
A model with predictions $d$ satisfies  $\epsilon-$Differential Equalized Odds ($\epsilon$-D) by verifying that for any $d$, $y$, $s_i$ and $s_j$, we have,

\begin{equation*}
 \begin{split}
 e^{-\epsilon} \le \frac{N_{d_1,y_k,s_i}}{N_{y_k,s_i}} \frac{N_{y_k,s_j}}{N_{d_1,y_k,s_j}} \le e^{\epsilon}
    \\
    \forall (s_i, s_j) \in A{\times}A, k \in \{0,1\}
\end{split}
\end{equation*}
Here, $d_1$ are positive predictions of a model, $y_k$, $k\in \{0,1\}$ are true labels, and  $s_i, s_j \in A$ are tuples of \textit{all} protected attributes (gender and racial/ethnic groups), including both intersectional subgroups (African American Woman) as well as general demographic groups (Woman).

\newpage

\section{Demographic Population Results}
\label{apd:pandemic-results}

\subsection{Depression Rates Before and During COVID-19 Pandemic}
\label{apd:pandemic-depression-results}
We compare the depression rates in the surveys submitted before and during the COVID-19 pandemic.
Each trial consisted in sampling surveys with replacement for each demographics and recorded the depression rates.
We performed 100 trials in order to get distributions of the dataset before and during the COVID-19 pandemic.
A pooled t-test revealed a significant difference in the depression rates of the surveys \textit{before} COVID-19 pandemic ($M=.56$, $SD=.028$) and \textit{during} COVID-19 pandemic ($M=0.83$, $SD=.015$),  $t(198)=83.9$, $p<.001$.

Looking at surveys submitted by men participants, a pooled t-test revealed a significant difference in the depression rates of the surveys \textit{before} COVID-19 pandemic ($M=.49$, $SD=.050$) and \textit{during} COVID-19 pandemic ($M=0.78$, $SD=.037$),  $t(198)=46.8$, $p<.001$.
Looking at surveys submitted by women participants, a pooled t-test revealed a significant difference in the depression rates of the surveys \textit{before} COVID-19 pandemic ($M=.59$, $SD=.032$) and \textit{during} COVID-19 pandemic ($M=0.84$, $SD=.017$),  $t(198)=67.5$, $p<.001$.

Looking at surveys submitted by White participants, a pooled t-test revealed a significant difference in the depression rates of the surveys \textit{before} COVID-19 pandemic ($M=.55$, $SD=.036$) and \textit{during} COVID-19 pandemic ($M=0.85$, $SD=.017$),  $t(198)=86.2$, $p<.001$.
Looking at surveys submitted by Black participants, a pooled t-test revealed a significant difference in the depression rates of the surveys \textit{before} COVID-19 pandemic ($M=.61$, $SD=.093$) and \textit{during} COVID-19 pandemic ($M=0.80$, $SD=.053$),  $t(198)=18.8$, $p<.001$.
Looking at surveys submitted by Hisp/Lat participants, a pooled t-test revealed a significant difference in the depression rates of the surveys \textit{before} COVID-19 pandemic ($M=.64$, $SD=.130$) and \textit{during} COVID-19 pandemic ($M=0.91$, $SD=.028$),  $t(198)=22.0$, $p<.001$.
Looking at surveys submitted by Asian participants, a pooled t-test revealed a significant difference in the depression rates of the surveys \textit{before} COVID-19 pandemic ($M=.47$, $SD=.099$) and \textit{during} COVID-19 pandemic ($M=0.66$, $SD=.064$),  $t(198)=16.7$, $p<.001$.

\subsection{Gender and Racial/Ethnic Group Rates Before and During the COVID-19 pandemic}
\label{apd:pandemic-demographic-results}
A $\chi^2$ test of independence was performed to examine the relation between gender and the number of surveys submitted before and during the COVID-19 pandemic. The relation between these variables was significant, $\chi^2$ (1, N = 2513) = 12.7, p = .00036. Women submitted more responses during the pandemic.
However, in the previous subsection we showed that the rates for depression where different before and during the pandemic, and given that we expect depression rates to be higher in women, perhaps the depressive symptom variable is a confounder.
We perform separate $\chi^2$ test of independence for each survey groups, with depressive symptoms and no depressive symptoms.
The relation between gender and the number of surveys with \textit{no depressive disorder} before and during the COVID-19 pandemic was \textit{not} significant $\chi^2$ (1, N = 1003) = 1.6, p = .213.
Also, the relation between gender and the number of surveys with \textit{Major or Minor depressive disorder} before and during the COVID-19 pandemic was \textit{not} significant $\chi^2$ (1, N = 1510) = 0.9, p = .366.
These results suggest that the gender populations did not change before and during the pandemic.

A $\chi^2$ test of independence was performed to examine the relation between race/ethnicity and the number of surveys submitted before and during the COVID-19 pandemic. The relation between these variables was significant, $\chi^2$ (4, N = 2607) = 81.2, $p < .001$. Asian American, African American and Hispanic/Latinx submitted more surveys during the pandemic.
This relation was significant for the surveys with \textit{no depressive disorder} $\chi^2$ (4, N = 1012) = 59.5, $p < .001$ and with \textit{Major or Minor depressive disorder} $\chi^2$ (4, N = 1595) = 36.4, $p < .001$.
These results suggest that Asian American, African American and Hispanic/Latinx submitted more surveys during the pandemic.
\newpage
\section{More Figures on Stressors}\label{apd:stressors-figures}

\begin{figure}[h]
\begin{minipage}{.45\linewidth}
\centering
  \includegraphics[width=\linewidth]{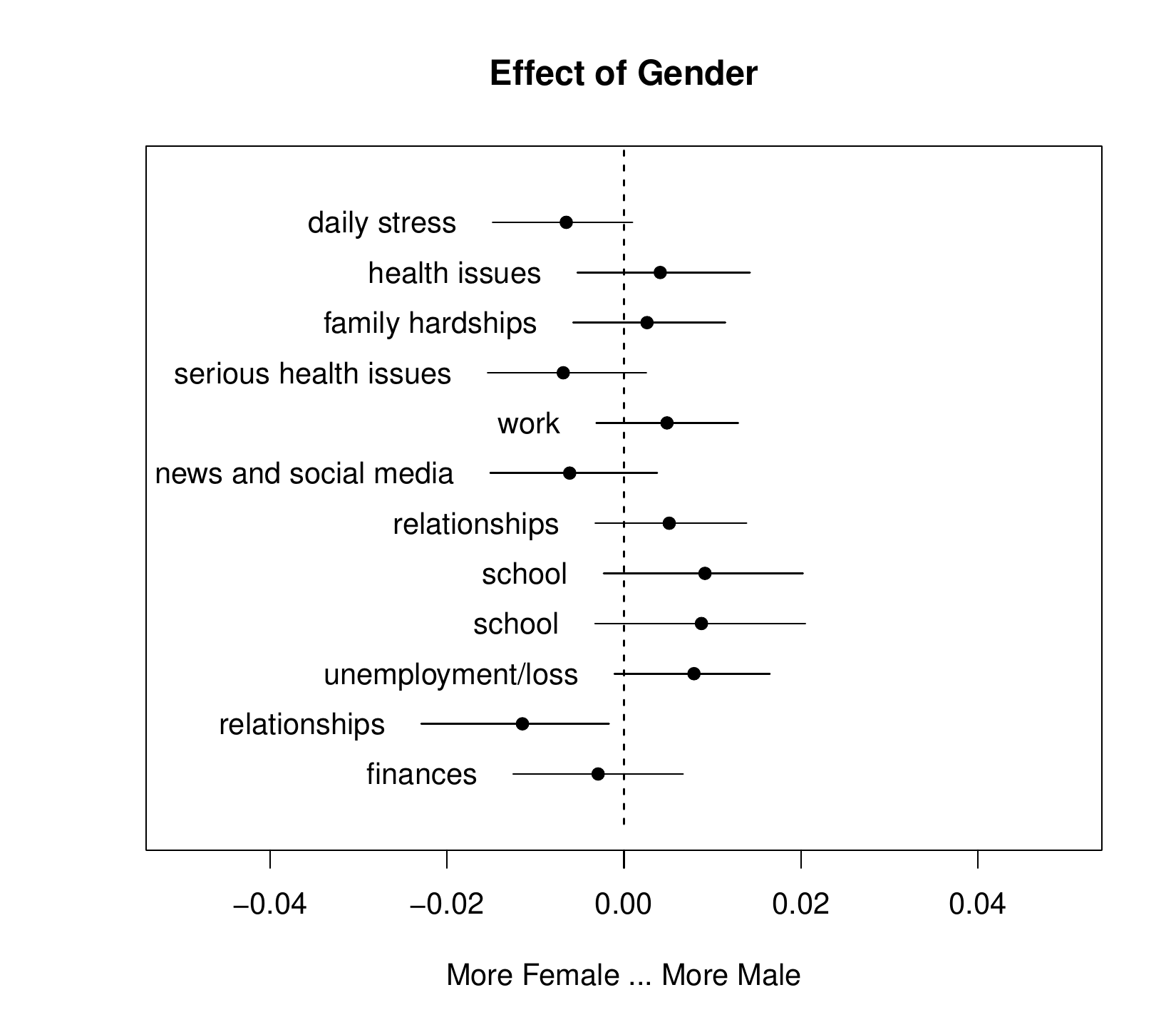}
  \caption{The effect of gender on topics about stressors.}
  \label{fig:topic-gender}
\end{minipage}
\hfill
\begin{minipage}{.45\linewidth}
\centering
  \includegraphics[width=\linewidth]{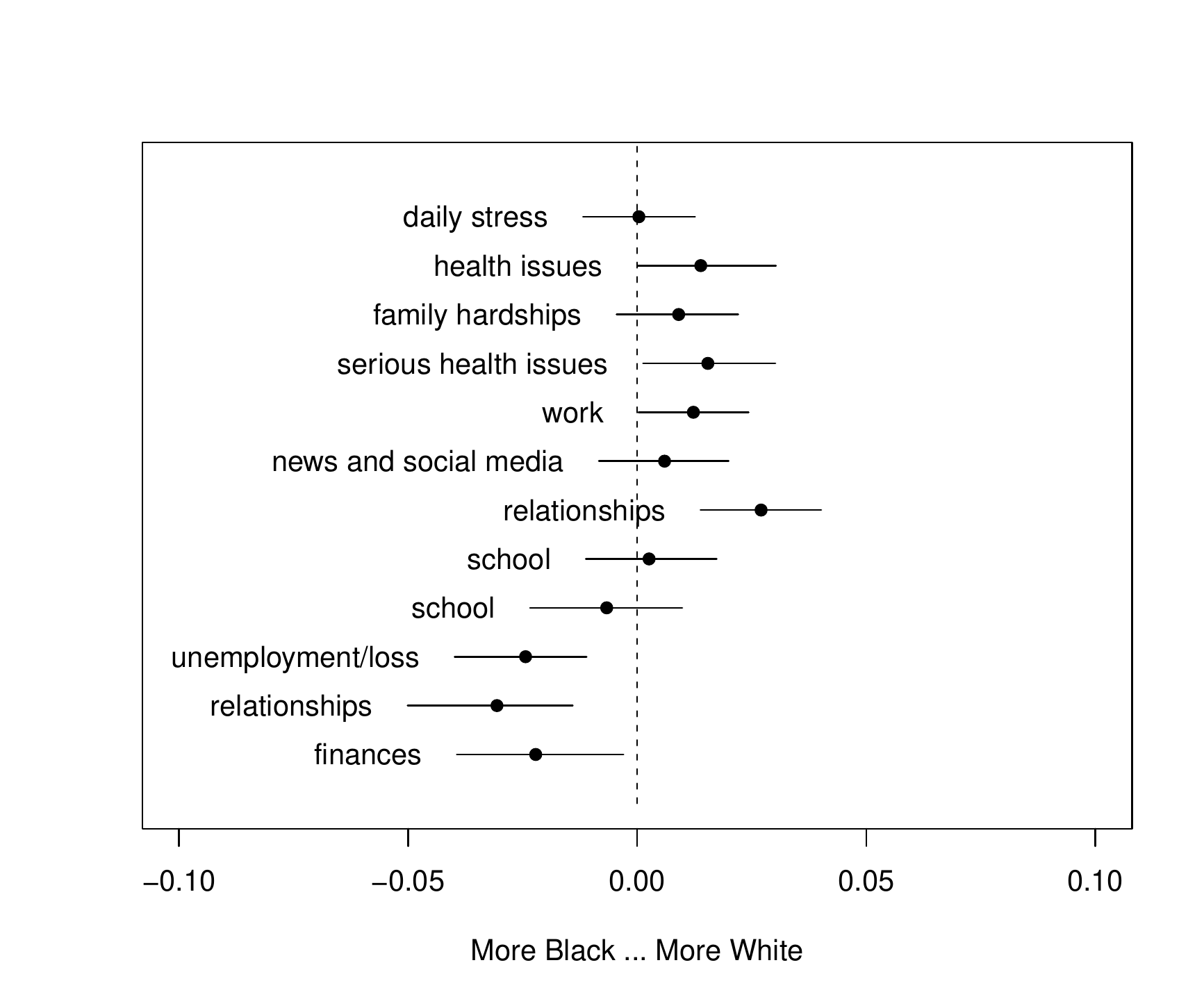}
  \caption{The effect of race (Black vs White population) on topics about stressors.}
  \label{fig:topic-race-white-black}
\end{minipage}
\end{figure}

\begin{figure}[h]
\begin{minipage}{.45\linewidth}
\centering
  \includegraphics[width=\linewidth]{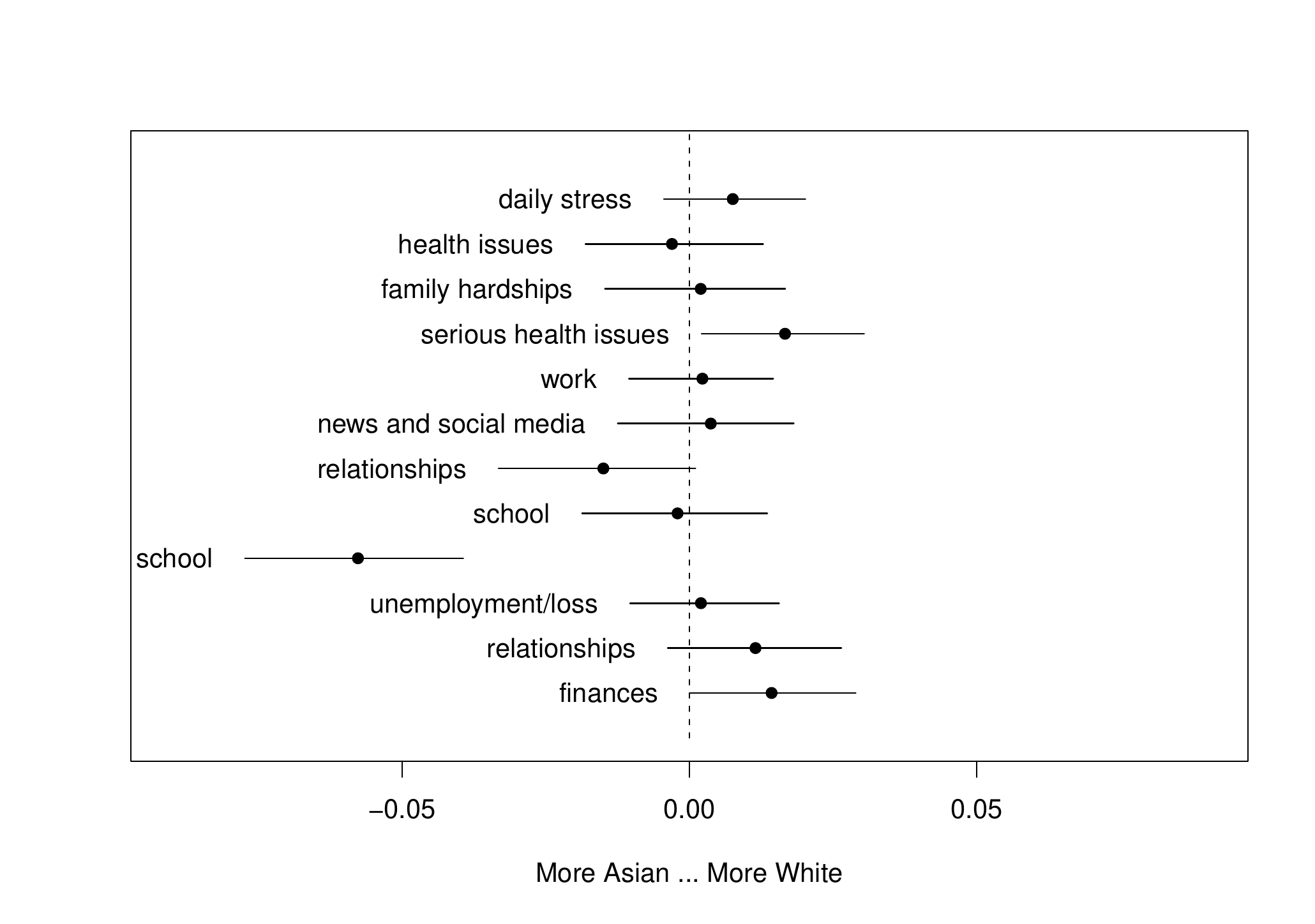}
  \caption{The effect of race (Asian vs White population) on topics about stressors.}
  \label{fig:topic-race-white-asian}
\end{minipage}
\hfill
\begin{minipage}{.45\linewidth}
\centering
  \includegraphics[width=\linewidth]{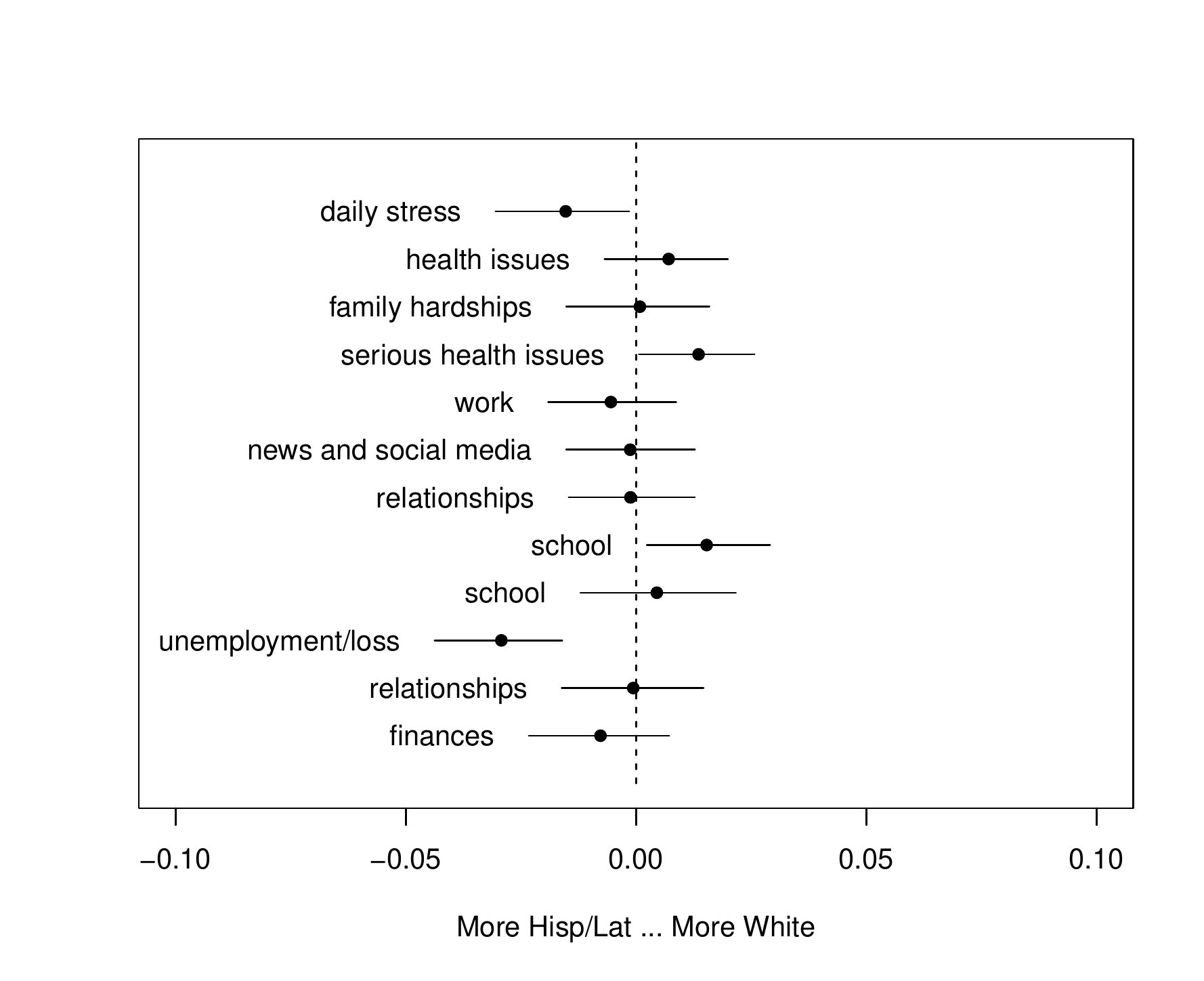}
  \caption{The effect of ethnicity (Hisp/Lat vs White population) on topics about stressors.}
  \label{fig:topic-race-white-hisp}
\end{minipage}
\end{figure}

\begin{figure}[h]
\begin{minipage}{.45\linewidth}
\centering
  \includegraphics[width=\linewidth]{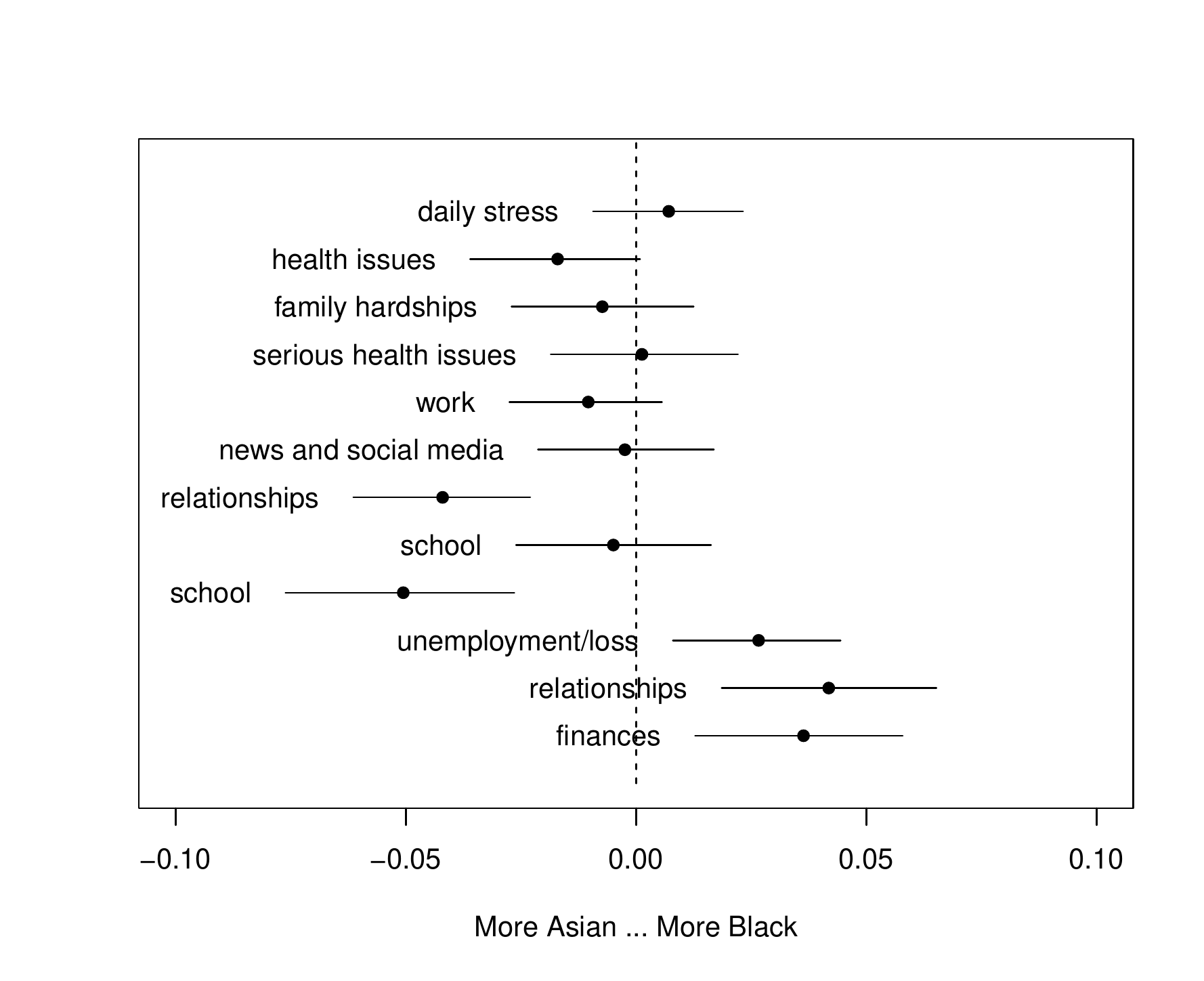}
  \caption{The effect of race (Asian vs Black population) on topics about stressors.}
  \label{fig:topic-race-black-asian}
\end{minipage}
\hfill
\begin{minipage}{.45\linewidth}
\centering
  \includegraphics[width=\linewidth]{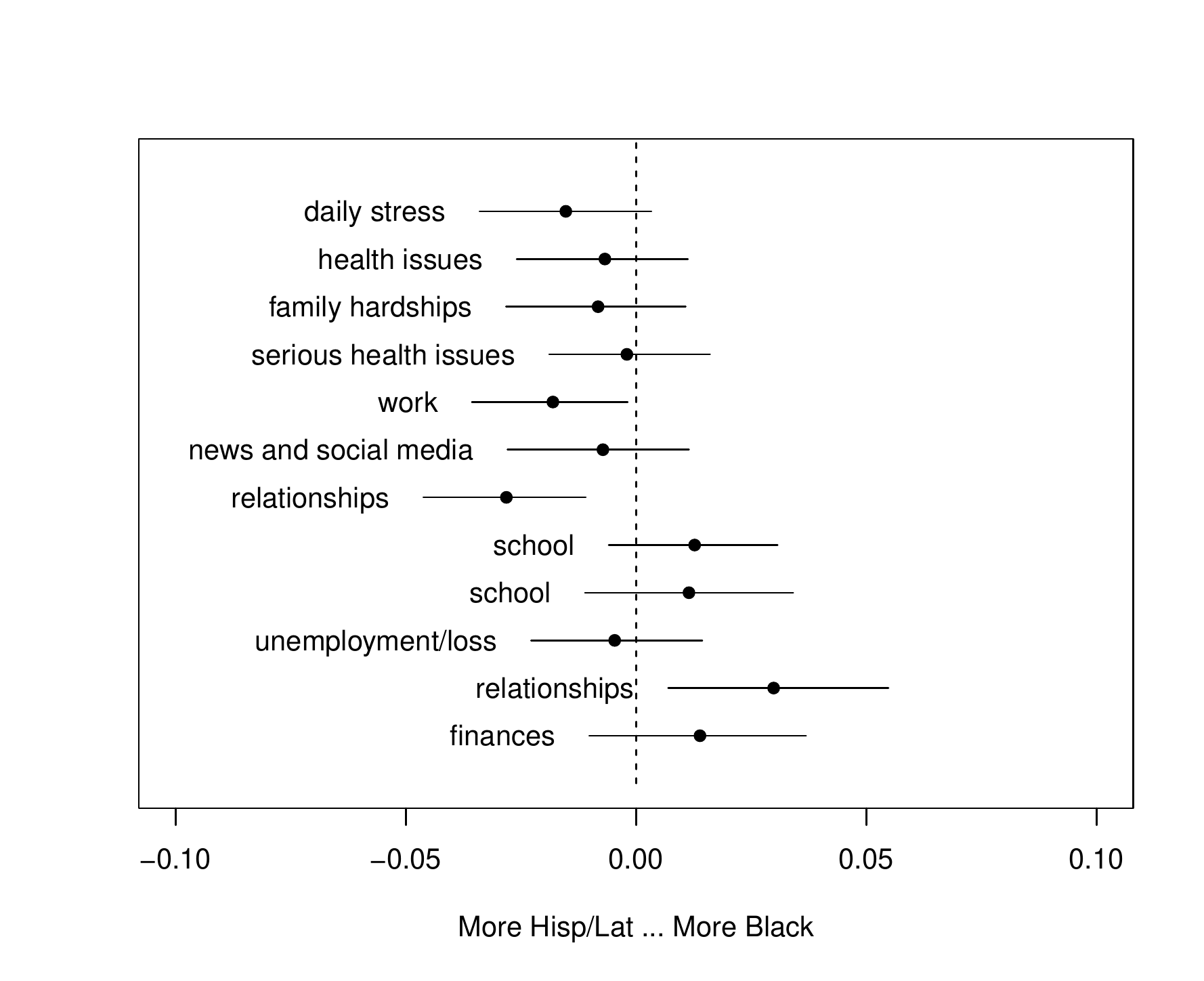}
  \caption{The effect of ethnicity (Hisp/Lat vs Black population) on topics about stressors.}
  \label{fig:topic-race-black-hisp}
\end{minipage}
\end{figure}

\begin{figure}[h]
\begin{minipage}{.45\linewidth}
\centering
  \includegraphics[width=\linewidth]{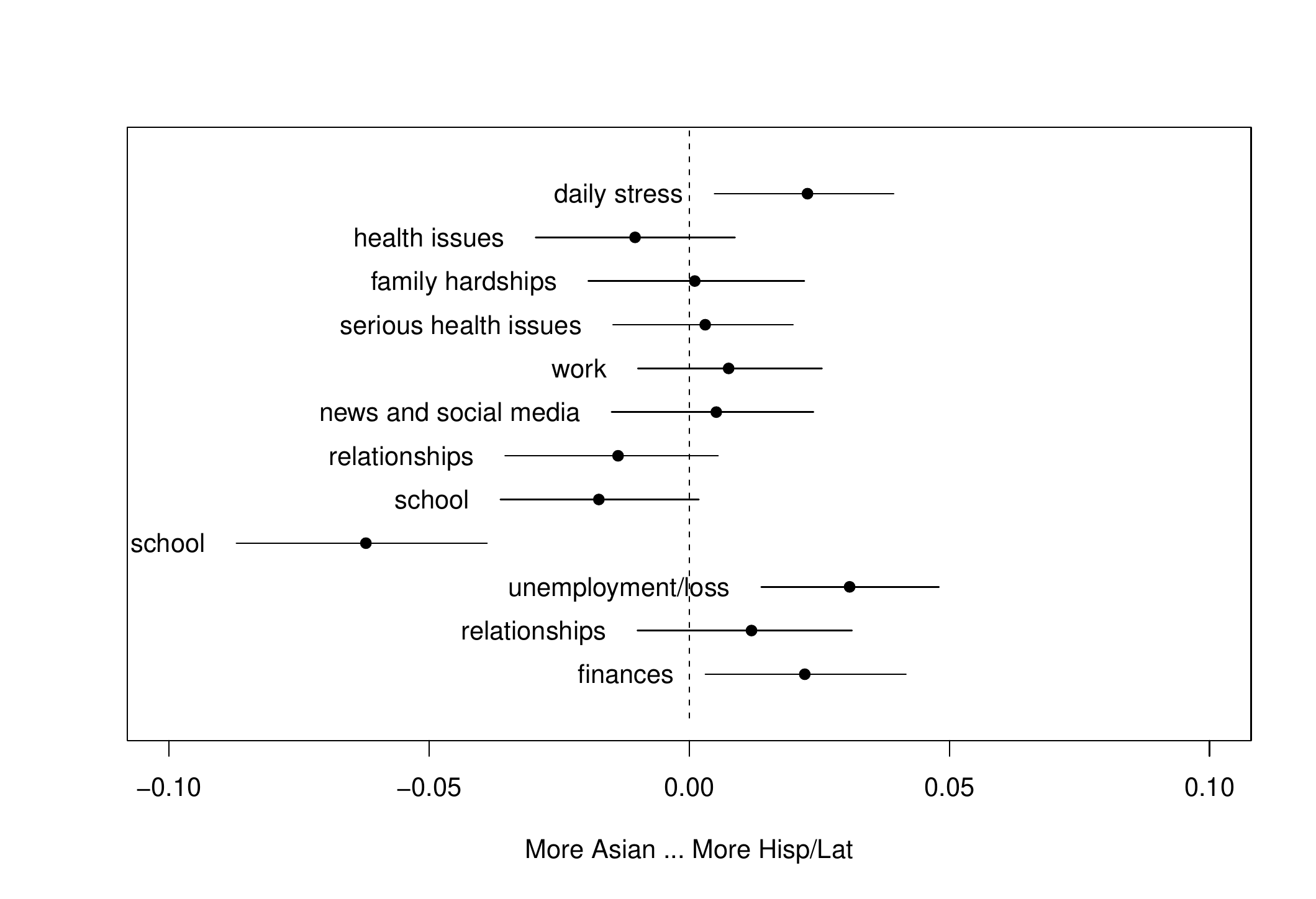}
  \caption{The effect of ethnicity (Asian vs Hisp/Lat population) on topics about stressors.}
  \label{fig:topic-race-hisp-asian}
\end{minipage}

\end{figure}

\tableref{tab:appx-vocab,tab:appx-liwc} show the pairwise log-odds-ratio comparison of token vocabulary and LIWC category distributions respectively.
\figureref{fig:topic-gender,fig:topic-race-white-black,fig:topic-race-white-asian,fig:topic-race-white-hisp,fig:topic-race-black-asian,fig:topic-race-black-hisp,fig:topic-race-hisp-asian} show the effects of gender and all racial/ethnic groups on the topics from the structural topic model trained on the stressors responses.

\begin{longtable}{llll}
\caption{Log-odds-ratio, Informative Dirichlet prior of the top 30 vocabulary tokens in all pairwise comparisons of the gender and racial/ethnic groups.} \label{tab:appx-vocab} \\

\toprule
\multicolumn{4}{c}{Gender}                                             \\
\midrule
\multicolumn{2}{c}{Women ($-$)}      & \multicolumn{2}{c}{Men ($+$)}   \\
\cmidrule(lr){1-2} \cmidrule(lr){3-4}
token               & ratio      & token               & ratio   \\
\cmidrule(lr){1-1} \cmidrule(lr){2-2} \cmidrule(lr){3-3} \cmidrule(lr){4-4}
n’t                & -4.501          & 's              & 0.053         \\
’m                 & -4.435          & homework        & 0.037         \\
’ve                & -3.621          & said            & 0.032         \\
’s                 & -3.225          & proactive       & 0.032         \\
covid              & -3.120          & finals          & 0.026         \\
covid-19           & -2.832          & consult         & 0.025         \\
’ll                & -1.562          & girlfriend      & 0.024         \\
coronavirus        & -0.948          & recovering      & 0.024         \\
’d                 & -0.800          & prescribed      & 0.023         \\
stress             & -0.483          & would           & 0.023         \\
’re                & -0.444          & relatives       & 0.022         \\
idk                & -0.377          & 're             & 0.022         \\
pandemic           & -0.314          & wife            & 0.021         \\
stressful          & -0.311          & deadlines       & 0.019         \\
alot               & -0.262          & first           & 0.019         \\
biggest            & -0.260          & may             & 0.019         \\
stressor           & -0.256          & looting         & 0.018         \\
trying             & -0.227          & entertained     & 0.018         \\
stressed           & -0.224          & might           & 0.018         \\
finances           & -0.212          & one             & 0.017         \\
stressing          & -0.200          & world           & 0.017         \\
source             & -0.191          & overcome        & 0.017         \\
cope               & -0.165          & dating          & 0.017         \\
schoolwork         & -0.164          & sexual          & 0.017         \\
exercising         & -0.158          & 'd              & 0.016         \\
boyfriend          & -0.151          & share           & 0.016         \\
dealing            & -0.137          & grades          & 0.016         \\
job                & -0.127          & contracting     & 0.016         \\
bills              & -0.126          & run             & 0.015         \\
gre                & -0.121          & could           & 0.015         \\
\midrule
\multicolumn{4}{c}{Ethnicity} \\
\midrule
\multicolumn{2}{c}{Black ($-$)}      & \multicolumn{2}{c}{White ($+$)} \\
\cmidrule(lr){1-2} \cmidrule(lr){3-4}
's                 & -0.069          & covid           & 4.798         \\
would              & -0.028          & n’t             & 4.494         \\
could              & -0.023          & ’ve             & 4.283         \\
n't                & -0.022          & ’m              & 4.227         \\
first              & -0.021          & covid-19        & 2.480         \\
might              & -0.019          & coronavirus     & 0.800         \\
two                & -0.018          & stress          & 0.523         \\
men                & -0.017          & pandemic        & 0.472         \\
mean               & -0.017          & stressful       & 0.335         \\
one                & -0.016          & stressor        & 0.284         \\
say                & -0.014          & stressed        & 0.260         \\
come               & -0.014          & dealing         & 0.236         \\
light              & -0.014          & biggest         & 0.233         \\
number             & -0.013          & finances        & 0.233         \\
great              & -0.013          & trying          & 0.224         \\
young              & -0.013          & source          & 0.183         \\
thank              & -0.013          & job             & 0.153         \\
morning            & -0.013          & work            & 0.140         \\
hopeful            & -0.012          & try             & 0.139         \\
across             & -0.012          & stressing       & 0.126         \\
programs           & -0.012          & relax           & 0.124         \\
hair               & -0.012          & cope            & 0.121         \\
front              & -0.012          & deal            & 0.120         \\
face               & -0.011          & worrying        & 0.118         \\
interest           & -0.011          & tried           & 0.112         \\
second             & -0.011          & applying        & 0.112         \\
brought            & -0.011          & yoga            & 0.111         \\
attention          & -0.011          & done            & 0.109         \\
available          & -0.011          & boyfriend       & 0.108         \\
george             & -0.011          & therapist       & 0.108         \\
\midrule
\multicolumn{2}{c}{Hisp/Lat ($-$)}   & \multicolumn{2}{c}{White ($+$)} \\
\cmidrule(lr){1-2} \cmidrule(lr){3-4}
idk                & -0.377          & covid           & 4.871         \\
's                 & -0.068          & n’t             & 4.459         \\
breaths            & -0.033          & ’ve             & 4.202         \\
're                & -0.029          & ’m              & 3.803         \\
would              & -0.026          & ’s              & 3.183         \\
could              & -0.023          & covid-19        & 2.773         \\
basic              & -0.022          & coronavirus     & 0.800         \\
n't                & -0.020          & stress          & 0.531         \\
first              & -0.019          & pandemic        & 0.332         \\
government         & -0.019          & stressful       & 0.323         \\
might              & -0.018          & stressor        & 0.256         \\
may                & -0.018          & biggest         & 0.244         \\
world              & -0.017          & dealing         & 0.231         \\
war                & -0.016          & trying          & 0.228         \\
one                & -0.016          & finances        & 0.222         \\
two                & -0.016          & stressed        & 0.215         \\
come               & -0.016          & source          & 0.190         \\
years              & -0.016          & internship      & 0.148         \\
light              & -0.015          & dealt           & 0.138         \\
ensure             & -0.015          & job             & 0.137         \\
store              & -0.015          & work            & 0.132         \\
lost               & -0.015          & relax           & 0.132         \\
name               & -0.014          & deal            & 0.129         \\
state              & -0.014          & stressing       & 0.126         \\
line               & -0.014          & try             & 0.122         \\
students           & -0.013          & cancelled       & 0.116         \\
eyes               & -0.013          & boyfriend       & 0.116         \\
say                & -0.013          & tried           & 0.115         \\
something          & -0.013          & applying        & 0.112         \\
idea               & -0.013          & therapist       & 0.108         \\
\midrule
\multicolumn{2}{c}{Asian ($-$)}      & \multicolumn{2}{c}{White ($+$)} \\
\cmidrule(lr){1-2} \cmidrule(lr){3-4}
's                 & -0.068          & covid           & 4.956         \\
said               & -0.040          & n’t             & 4.459         \\
scheduling         & -0.030          & ’ve             & 4.202         \\
internships        & -0.030          & ’m              & 4.039         \\
schoolwork         & -0.029          & ’s              & 3.022         \\
would              & -0.026          & covid-19        & 2.480         \\
first              & -0.020          & ’ll             & 1.753         \\
n't                & -0.019          & ’re             & 0.800         \\
black              & -0.018          & coronavirus     & 0.718         \\
might              & -0.018          & stress          & 0.543         \\
insecurity         & -0.017          & pandemic        & 0.437         \\
two                & -0.017          & stressful       & 0.299         \\
may                & -0.017          & biggest         & 0.262         \\
years              & -0.016          & stressed        & 0.251         \\
one                & -0.016          & dealing         & 0.245         \\
state              & -0.015          & finances        & 0.233         \\
looked             & -0.015          & trying          & 0.220         \\
history            & -0.014          & source          & 0.200         \\
however            & -0.014          & destress        & 0.174         \\
went               & -0.014          & job             & 0.153         \\
students           & -0.014          & dealt           & 0.145         \\
come               & -0.014          & deal            & 0.129         \\
goes               & -0.013          & internship      & 0.129         \\
think              & -0.013          & work            & 0.126         \\
young              & -0.013          & exercising      & 0.121         \\
financially        & -0.013          & getting         & 0.117         \\
play               & -0.012          & relax           & 0.116         \\
three              & -0.012          & cancelled       & 0.116         \\
market             & -0.012          & boyfriend       & 0.116         \\
chat               & -0.012          & tried           & 0.112         \\
\midrule
\multicolumn{2}{c}{Hisp/Lat ($-$)}   & \multicolumn{2}{c}{Black ($+$)} \\
\cmidrule(lr){1-2} \cmidrule(lr){3-4}
’m                 & -1.349          & covid-19        & 0.800         \\
’ve                & -0.628          & covid           & 0.242         \\
n’t                & -0.628          & finding         & 0.015         \\
pandemic           & -0.141          & exams           & 0.013         \\
stressed           & -0.045          & moving          & 0.012         \\
lost               & -0.030          & studying        & 0.011         \\
stressor           & -0.029          & anxiety         & 0.011         \\
bills              & -0.025          & biggest         & 0.011         \\
virus              & -0.024          & deal            & 0.010         \\
due                & -0.023          & paying          & 0.009         \\
daily              & -0.022          & housing         & 0.009         \\
try                & -0.018          & currently       & 0.009         \\
basic              & -0.017          & money           & 0.009         \\
job                & -0.016          & knowing         & 0.008         \\
feel               & -0.015          & stress          & 0.008         \\
mask               & -0.014          & diagnosed       & 0.008         \\
hopeless           & -0.014          & relax           & 0.008         \\
done               & -0.014          & mental          & 0.008         \\
sleep              & -0.013          & anxious         & 0.008         \\
getting            & -0.013          & save            & 0.007         \\
stressful          & -0.012          & boyfriend       & 0.007         \\
struggling         & -0.011          & source          & 0.007         \\
mood               & -0.011          & lack            & 0.007         \\
working            & -0.011          & care            & 0.006         \\
family             & -0.011          & weeks           & 0.005         \\
finances           & -0.011          & neighbors       & 0.005         \\
unemployed         & -0.010          & right           & 0.005         \\
hard               & -0.010          & busy            & 0.005         \\
walks              & -0.010          & move            & 0.005         \\
current            & -0.010          & talking         & 0.005         \\
\midrule
\multicolumn{2}{c}{Asian ($-$)}      & \multicolumn{2}{c}{Black ($+$)} \\
\cmidrule(lr){1-2} \cmidrule(lr){3-4}
’m                 & -0.893          & covid           & 1.054         \\
’ve                & -0.628          & biggest         & 0.029         \\
n’t                & -0.628          & paying          & 0.022         \\
coronavirus        & -0.082          & stress          & 0.021         \\
assignments        & -0.038          & currently       & 0.017         \\
stressful          & -0.036          & source          & 0.017         \\
pandemic           & -0.035          & saving          & 0.013         \\
applying           & -0.033          & getting         & 0.012         \\
try                & -0.030          & finding         & 0.011         \\
schedule           & -0.026          & life            & 0.010         \\
friends            & -0.022          & lack            & 0.010         \\
managing           & -0.019          & dealing         & 0.010         \\
stressing          & -0.018          & deal            & 0.009         \\
cope               & -0.018          & moving          & 0.009         \\
grad               & -0.017          & car             & 0.008         \\
school             & -0.016          & mental          & 0.008         \\
things             & -0.016          & sources         & 0.008         \\
taking             & -0.015          & boyfriend       & 0.007         \\
organized          & -0.015          & safe            & 0.007         \\
work               & -0.014          & doctor          & 0.007         \\
exams              & -0.013          & weight          & 0.007         \\
time               & -0.013          & relatives       & 0.006         \\
thinking           & -0.012          & unemployment    & 0.006         \\
meditation         & -0.012          & grades          & 0.006         \\
stresses           & -0.011          & stable          & 0.006         \\
week               & -0.010          & feeling         & 0.006         \\
feel               & -0.010          & financial       & 0.006         \\
financially        & -0.010          & pregnant        & 0.006         \\
talked             & -0.010          & studying        & 0.006         \\
still              & -0.010          & constantly      & 0.006         \\
\midrule
\multicolumn{2}{l}{Hisp/Lat ($-$)} & \multicolumn{2}{l}{Asian ($+$) }   \\
\cmidrule(lr){1-2} \cmidrule(lr){3-4}
covid              & -0.824          & covid-19        & 0.800         \\
’m                 & -0.499          & ’s              & 0.258         \\
pandemic           & -0.106          & schoolwork      & 0.110         \\
exercising         & -0.037          & coronavirus     & 0.082         \\
stressed           & -0.036          & assignments     & 0.039         \\
distancing         & -0.025          & deadlines       & 0.036         \\
getting            & -0.024          & applying        & 0.033         \\
daily              & -0.023          & schedule        & 0.026         \\
masks              & -0.020          & exams           & 0.026         \\
due                & -0.020          & concentrate     & 0.025         \\
biggest            & -0.018          & stressful       & 0.024         \\
thoughts           & -0.018          & friends         & 0.022         \\
job                & -0.017          & managing        & 0.019         \\
sleep              & -0.017          & motivation      & 0.019         \\
suicidal           & -0.014          & internship      & 0.019         \\
dealing            & -0.014          & stressing       & 0.018         \\
paying             & -0.013          & school          & 0.017         \\
making             & -0.013          & relax           & 0.016         \\
hard               & -0.013          & try             & 0.012         \\
current            & -0.013          & stresses        & 0.011         \\
stress             & -0.012          & balancing       & 0.010         \\
car                & -0.012          & time            & 0.010         \\
needs              & -0.011          & financially     & 0.010         \\
mood               & -0.011          & jobs            & 0.009         \\
feeling            & -0.011          & keep            & 0.009         \\
constantly         & -0.011          & talking         & 0.009         \\
ca                 & -0.011          & things          & 0.008         \\
finances           & -0.011          & knowing         & 0.008         \\
walks              & -0.010          & awake           & 0.008         \\
'm                 & -0.010          & lonely          & 0.008         \\
\bottomrule

\end{longtable}

\begin{longtable}{llll}
\caption{Log-odds-ratio, Informative Dirichlet prior of the top 10 LIWC categories in all pairwise comparisons of the gender and racial/ethnic groups.} \label{tab:appx-liwc} \\

\toprule
\multicolumn{4}{c}{Gender}                                             \\
\midrule
\multicolumn{2}{c}{Women ($-$)}      & \multicolumn{2}{c}{Men ($+$)}   \\
\cmidrule(lr){1-2} \cmidrule(lr){3-4}
category               & ratio      & category               & ratio   \\
\cmidrule(lr){1-1} \cmidrule(lr){2-2} \cmidrule(lr){3-3} \cmidrule(lr){4-4}
anx                 & -0.184       & article            & 0.130      \\
negemo              & -0.119       & function           & 0.125      \\
verb                & -0.110       & prep               & 0.077      \\
pro1                & -0.095       & social             & 0.071      \\
focuspresent        & -0.095       & shehe              & 0.060      \\
achiev              & -0.094       & conj               & 0.041      \\
health              & -0.093       & hear               & 0.041      \\
affect              & -0.072       & number             & 0.039      \\
drives              & -0.066       & space              & 0.038      \\
sad                 & -0.062       & female             & 0.037      \\
family              & -0.061       & pronoun            & 0.036      \\
money               & -0.061       & percept            & 0.036      \\
reward              & -0.059       & male               & 0.034      \\
home                & -0.050       & ipron              & 0.030      \\
cogproc             & -0.042       & see                & 0.027      \\
\midrule
\multicolumn{4}{c}{Ethnicity}                                        \\
\midrule
\multicolumn{2}{c}{Black ($-$)}    & \multicolumn{2}{c}{White ($+$)} \\
\cmidrule(lr){1-2} \cmidrule(lr){3-4}
article             & -0.164       & anx                & 0.197      \\
function            & -0.153       & health             & 0.131      \\
prep                & -0.084       & negemo             & 0.128      \\
social              & -0.080       & verb               & 0.126      \\
shehe               & -0.077       & achiev             & 0.119      \\
conj                & -0.056       & focuspresent       & 0.114      \\
male                & -0.054       & pro1               & 0.110      \\
space               & -0.049       & affect             & 0.086      \\
hear                & -0.049       & family             & 0.080      \\
pronoun             & -0.047       & drives             & 0.076      \\
number              & -0.046       & sad                & 0.069      \\
percept             & -0.044       & time               & 0.064      \\
power               & -0.038       & friend             & 0.060      \\
ipron               & -0.038       & bio                & 0.058      \\
female              & -0.036       & home               & 0.053      \\
\midrule
\multicolumn{2}{c}{Hisp/Lat ($-$)} & \multicolumn{2}{c}{White ($+$)} \\
\cmidrule(lr){1-2} \cmidrule(lr){3-4}
article             & -0.156       & anx                & 0.193      \\
function            & -0.146       & health             & 0.133      \\
prep                & -0.077       & verb               & 0.121      \\
social              & -0.077       & negemo             & 0.119      \\
shehe               & -0.072       & achiev             & 0.111      \\
conj                & -0.053       & focuspresent       & 0.108      \\
male                & -0.049       & pro1               & 0.103      \\
hear                & -0.048       & affect             & 0.083      \\
percept             & -0.046       & family             & 0.076      \\
pronoun             & -0.046       & drives             & 0.070      \\
space               & -0.045       & time               & 0.064      \\
number              & -0.043       & friend             & 0.062      \\
power               & -0.039       & sad                & 0.059      \\
ipron               & -0.038       & bio                & 0.056      \\
female              & -0.035       & reward             & 0.051      \\
\midrule
\multicolumn{2}{c}{Asian ($-$)}    & \multicolumn{2}{c}{White ($+$)} \\
\cmidrule(lr){1-2} \cmidrule(lr){3-4}
article             & -0.161       & anx                & 0.204      \\
function            & -0.150       & health             & 0.141      \\
prep                & -0.083       & negemo             & 0.133      \\
social              & -0.078       & verb               & 0.123      \\
shehe               & -0.073       & pro1               & 0.110      \\
conj                & -0.057       & achiev             & 0.107      \\
hear                & -0.050       & focuspresent       & 0.106      \\
male                & -0.050       & affect             & 0.090      \\
percept             & -0.046       & family             & 0.081      \\
space               & -0.045       & sad                & 0.073      \\
number              & -0.045       & drives             & 0.073      \\
pronoun             & -0.045       & bio                & 0.066      \\
ipron               & -0.039       & time               & 0.056      \\
power               & -0.035       & friend             & 0.055      \\
see                 & -0.034       & reward             & 0.053      \\
\midrule
\multicolumn{2}{c}{Hisp/Lat ($-$)} & \multicolumn{2}{c}{Black ($+$)} \\
\cmidrule(lr){1-2} \cmidrule(lr){3-4}
sad                 & -0.009       & article            & 0.008      \\
negemo              & -0.008       & prep               & 0.007      \\
achiev              & -0.008       & function           & 0.006      \\
focuspresent        & -0.007       & focusfuture        & 0.004      \\
pro1                & -0.007       & shehe              & 0.004      \\
adverb              & -0.006       & relig              & 0.004      \\
verb                & -0.006       & male               & 0.004      \\
anx                 & -0.006       & space              & 0.004      \\
feel                & -0.005       & social             & 0.003      \\
home                & -0.005       & relativ            & 0.003      \\
informal            & -0.005       & number             & 0.003      \\
body                & -0.005       & conj               & 0.003      \\
leisure             & -0.004       & motion             & 0.003      \\
affect              & -0.004       & interrog           & 0.003      \\
tentat              & -0.004       & risk               & 0.003      \\
\midrule
\multicolumn{2}{c}{Asian ($-$)}    & \multicolumn{2}{c}{Black ($+$)} \\
\cmidrule(lr){1-2} \cmidrule(lr){3-4}
work                & -0.013       & health             & 0.009      \\
achiev              & -0.011       & bio                & 0.008      \\
leisure             & -0.010       & negemo             & 0.005      \\
focuspresent        & -0.009       & anx                & 0.004      \\
time                & -0.007       & risk               & 0.004      \\
friend              & -0.005       & motion             & 0.004      \\
affiliation         & -0.004       & money              & 0.004      \\
home                & -0.004       & male               & 0.003      \\
verb                & -0.003       & shehe              & 0.003      \\
see                 & -0.003       & article            & 0.003      \\
drives              & -0.003       & ingest             & 0.003      \\
informal            & -0.002       & female             & 0.003      \\
ipron               & -0.002       & affect             & 0.003      \\
compare             & -0.002       & ppron              & 0.003      \\
percept             & -0.002       & power              & 0.003      \\
\midrule
\multicolumn{2}{c}{Hisp/Lat ($-$)} & \multicolumn{2}{c}{Asian ($+$)} \\
\cmidrule(lr){1-2} \cmidrule(lr){3-4}
negemo              & -0.013       & work               & 0.015      \\
sad                 & -0.013       & time               & 0.008      \\
anx                 & -0.010       & friend             & 0.006      \\
health              & -0.010       & relativ            & 0.006      \\
bio                 & -0.010       & prep               & 0.005      \\
pro1                & -0.007       & leisure            & 0.005      \\
affect              & -0.007       & article            & 0.005      \\
feel                & -0.006       & focusfuture        & 0.004      \\
body                & -0.006       & function           & 0.004      \\
family              & -0.006       & conj               & 0.003      \\
money               & -0.005       & compare            & 0.003      \\
certain             & -0.005       & achiev             & 0.003      \\
adverb              & -0.005       & focuspresent       & 0.003      \\
cogproc             & -0.005       & number             & 0.002      \\
risk                & -0.003       & hear               & 0.002      \\
\bottomrule
\end{longtable}

\end{document}